\pdfoutput=1

\documentclass[11pt]{article}

\usepackage[]{acl}

\usepackage{times}
\usepackage{latexsym}

\usepackage[T1]{fontenc}

\usepackage[utf8]{inputenc}

\usepackage{microtype}

\usepackage{inconsolata}

\usepackage{graphicx}
\usepackage{booktabs}
\usepackage{multicol}
\usepackage{multirow}
\usepackage{amsmath}

\title{Document-level Causal Relation Extraction with \\ Knowledge-guided Binary Question Answering}

\author{
  Zimu Wang, Lei Xia, Wei Wang, Xinya Du \\
  University of Texas at Dallas \\
  \texttt{\{zimu.wang, lei.xia, xinya.du\}@utdallas.edu} \\
}

\begin{document}
\maketitle
\begin{abstract}
As an essential task in information extraction (IE), Event-Event Causal Relation Extraction (ECRE) aims to identify and classify the causal relationships between event mentions in natural language texts. However, existing research on ECRE has highlighted two critical challenges, including the lack of document-level modeling and causal hallucinations.
In this paper, we propose a \textbf{Know}ledge-guided binary \textbf{Q}uestion \textbf{A}nswering (\textbf{KnowQA}) method with event structures for ECRE, consisting of two stages: \textit{Event Structure Construction} and \textit{Binary Question Answering}.
We conduct extensive experiments under both zero-shot and fine-tuning settings with large language models (LLMs) on the MECI and MAVEN-ERE datasets. Experimental results demonstrate the usefulness of event structures on document-level ECRE and the effectiveness of KnowQA by achieving state-of-the-art on the MECI dataset.
We observe not only the effectiveness but also the high generalizability and low inconsistency of our method, particularly when with complete event structures after fine-tuning the models\footnote{The source code for this paper is publicly released at \url{https://github.com/du-nlp-lab/KnowQA}.}.
\end{abstract}

\section{Introduction}

\begin{figure}[t!]
    \centering
    \includegraphics[width=\linewidth]{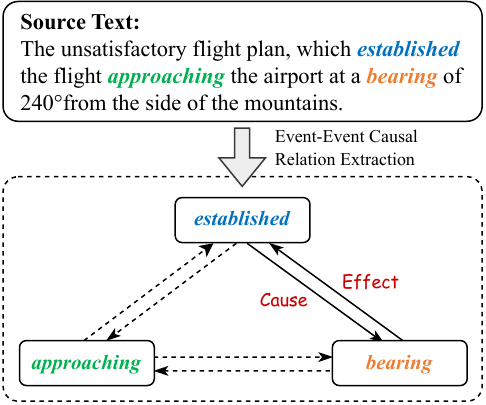}
    \caption{Overview of the ECRE process. The dashed lines indicate there are no causal relationships between the event mentions.}
    \label{fig:eci-intro}
\end{figure}

Event-Event Causal Relation Extraction (ECRE) is an essential task in information extraction (IE) that aims to identify and classify the causal relationships between event mentions in natural language texts. For example, given a sentence and an event mention pair of interest (\textbf{\textit{established}}, \textbf{\textit{bearing}}), an ECRE model should recognize the causal relationship between them, i.e., \textbf{\textit{established}} $\xrightarrow{cause}$ \textbf{\textit{bearing}} (Figure \ref{fig:eci-intro}). ECRE is regarded as a precondition of various downstream tasks, such as event knowledge graph construction \cite{ma-etal-2022-mmekg}, future event prediction \cite{10.1145/3477495.3532080}, machine reading comprehension \cite{zhu-etal-2023-causal}, and natural language logical/temporal reasoning \cite{yang-etal-2020-improving,yang-etal-2023-end,yang-etal-2024-language}.

Early studies of ECRE mainly focus on identifying the existence of causal relationships, regarding it as a binary classification task and ignoring the directions of these relationships. Researchers have utilized pre-trained language models (PLMs) to model the contexts and improved performance by enriching the event representations and modeling event associations \cite{tran-phu-nguyen-2021-graph,hu-etal-2023-semantic}. Some research has also investigated the potency of large language models (LLMs) on this task \cite{gao-etal-2023-chatgpt}. Recently, the availability of large-scale datasets makes it possible to classify the causal relationships between event mentions \cite{lai-etal-2022-meci,wang-etal-2022-maven}, which is much more challenging since it necessitates fully understanding the contexts and determining the cause and effect of each event pair while likewise taking additional factors like language varieties into account. In general, existing research on ECRE has highlighted the following two critical challenges:

(1) \textbf{Lack of Document-level Modeling.} Existing ECRE models typically leverage semantic structures, particularly the Abstract Meaning Representation (AMR) graph \cite{banarescu-etal-2013-abstract}, to model event-related contextual information, where the nodes of the graphs represent events, entities, etc., and the edges denote the semantic relationships between them. However, as AMR graphs are built at the sentence level, they are limited in capturing document-level semantics, restricting their capacity to identify implicit and fine-grained information in texts \cite{tran-phu-nguyen-2021-graph,hu-etal-2023-semantic}. Consequently, some approaches that apply AMR graphs for document modeling have not been evaluated for their performance in inter-sentence ECRE \cite{hu-etal-2023-semantic}.

(2) \textbf{Causal Hallucinations.} LLMs, such as ChatGPT and GPT-4, have been shown to fall short on the ECRE task and suffer from serious causal hallucination issues by overestimating the existence of causal relationships, largely attributed to reporting biases in natural language, where causal relationships are often described, while the events involved in these relationships are not expressed explicitly \cite{gao-etal-2023-chatgpt}. This phenomenon contributes to low precision and high recall of LLMs on this task, which severely hampers their performance in this field \cite{gao-etal-2023-chatgpt,liu-etal-2024-identifying}.
Such challenges are critical to designing more reliable ECRE models, particularly at the document level with appropriate document-level semantic features.

In this paper, we propose a \textbf{Know}ledge-guided binary \textbf{Q}uestion \textbf{A}nswering (\textbf{KnowQA}) method to deal with the aforementioned challenges. Unlike previous work that relys heavily on semantic structures, we leverage cross-task knowledge to construct document-level event structures to enrich event information, motivated by the effectiveness of cross-task knowledge in IE \cite{lin-etal-2020-joint,jin-etal-2023-toward}. We define the ECRE task as consisting of the two subtasks: (1) \textbf{Event Causality Identification (ECI)}, which identifies the existence of causal relationships, and (2) \textbf{Causal Relation Classification (CRC)}, which classifies the event pairs containing causal relationships into their corresponding relation types.
As shown in Figure \ref{fig:framework}, the overall framework of KnowQA consists of two stages: \textit{Event Structure Construction} and \textit{Binary Question Answering}. In the first stage, we extend the event structures as event mentions, event arguments, and the single-hop relationships of arguments, and we utilize IE models to construct them at the document level. Then, we formulate ECRE as a binary question answering (QA) task with single-turn (for identification) and multi-turn (for identification and classification) strategies with specific relation types, and we incorporate the constructed event structures into the questions.

We conduct comprehensive experiments under zero-shot and fine-tuning settings with LLMs. Experimental results on the MECI \cite{lai-etal-2022-meci} and MAVEN-ERE \cite{wang-etal-2022-maven} datasets demonstrate the effectiveness of KnowQA by outperforming the baseline models and achieving the state-of-the-art on the MECI dataset. We also discuss the following benefits of our approach brings:
(1) the effectiveness of complete event structures in the ECRE task, particularly at the document level;
(2) the efficacy of multi-turn QA under the zero-shot setting and single-turn QA for the identification and multi-turn QA for the classification of causal relationships under the fine-tuning setting; and 
(3) the high generalizability and low inconsistency of our multi-turn QA strategy, particularly when with complete event structures after fine-tuning the models.

The key contributions of this work are summarized as follows:

\begin{figure*}[t!]
    \centering
    \includegraphics[width=1\linewidth]{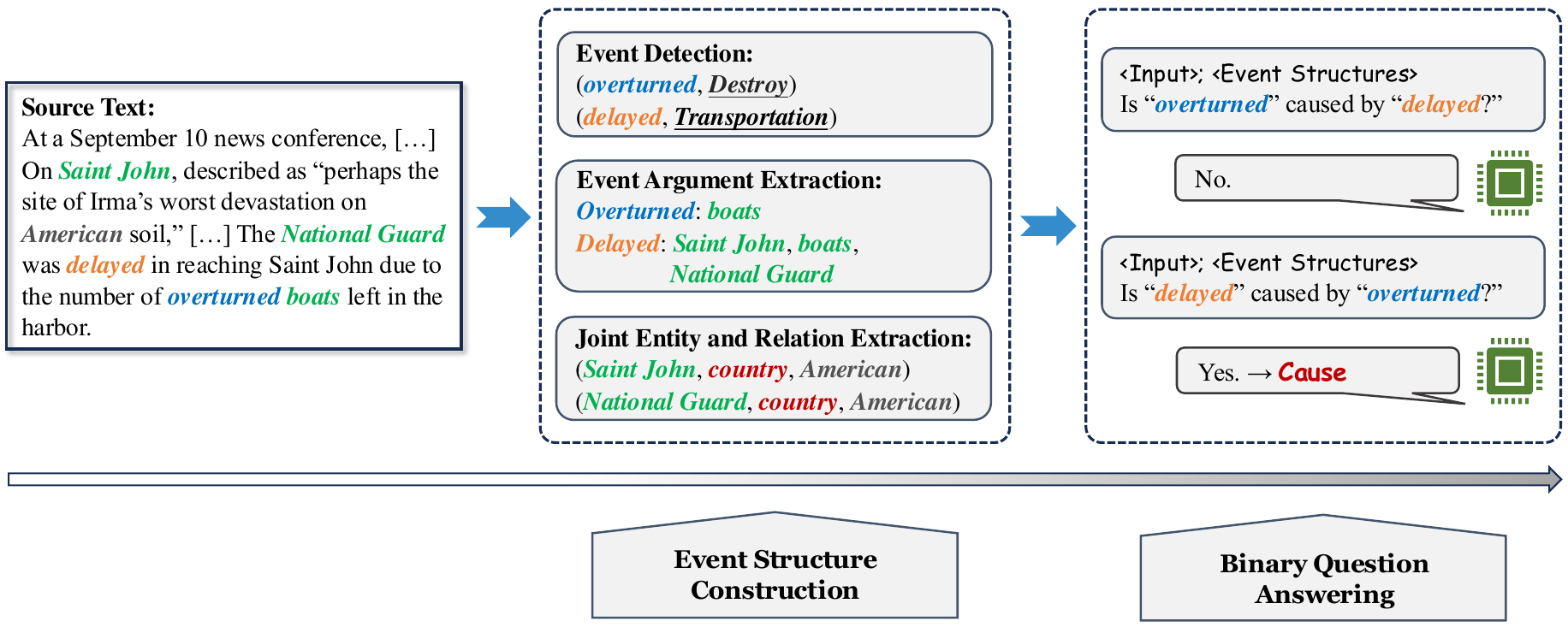}
    \caption{Overall framework of the proposed KnowQA method for ECRE, consisting of two stages: \textit{Event Structure Construction} and \textit{Binary Question Answering}. In the first stage, we utilize IE models to form event structures at the document level. In the second stage, we formulate ECRE as a binary QA task with single-turn and multi-turn strategies and fully leverage the event structures for ECRE predictions.}
    \label{fig:framework}
\end{figure*}

\begin{itemize}
    \item We propose KnowQA, formulating ECRE as a binary QA task with single-turn and multi-turn strategies. To the best of our knowledge, we are the first to utilize QA strategies for ECRE with specific relation types.
    \item We extend the event structures as event mentions, event arguments, and the single-hop relationships of arguments and validate their effectiveness in the ECRE task.
    \item We demonstrate the effectiveness of KnowQA, particularly at the document level, and we discuss the high generalizability and low inconsistency of our method.
\end{itemize}

\section{Related Work}

\paragraph{Event-Event Causal Relation Extraction.}

The field of ECRE has been increasingly recognized for its diverse applications; however, research on ECRE has mainly focused on the ECI task that ignores the directions of causal relationships. Early studies on ECI have concentrated on utilizing syntactic patterns \cite{riaz-girju-2013-toward,gao-etal-2019-modeling}, statistical event occurrences \cite{do-etal-2011-minimally,hu-walker-2017-inferring}, and weakly supervised data \cite{hashimoto-2019-weakly}. Additionally, recent advancements have leveraged PLMs and introduce semantic structures \cite{tran-phu-nguyen-2021-graph,hu-etal-2023-semantic}, external knowledge \cite{ijcai2020p499,cao-etal-2021-knowledge}, and data augmentation \cite{zuo-etal-2020-knowdis,zuo-etal-2021-learnda} approaches and investigated the potency of ECI with LLMs \cite{gao-etal-2023-chatgpt}. Recently, with the availability of large-scale datasets, some research also focuses on the classification of causal relationships to their corresponding relation types \cite{deng-etal-2023-speech,hu2023protoemprototypeenhancedmatchingframework}.

However, previous research has struggled to model document-level event-related contextual information comprehensively, as the semantic structures are typically confined to the sentence level. In this paper, we construct document-level event structures and enrich them with event mentions for ECRE predictions.

\paragraph{QA-based IE.}

Motivated by the effectiveness of natural languages in offering external supervision and mitigating the gap between contexts and tasks, QA-based methods have been extensively studied in the field of IE.
\newcite{du-cardie-2020-event, liu-etal-2020-event} utilize heuristic-based methods for generating questions.
\newcite{du-ji-2022-retrieval,choudhary-du-2024-qaevent} leverage end-to-end deep learning-based methods \cite{du-etal-2017-learning,du-cardie-2018-harvesting} for generating QA pairs for representing events.
In addition, QA-based methods have also been investigated in temporal relation extraction \cite{cohen-bar-2023-temporal} and ECI \cite{gao-etal-2023-chatgpt}, or retrieving useful background knowledge to improve event causality recognition \cite{Kruengkrai_Torisawa_Hashimoto_Kloetzer_Oh_Tanaka_2017,kadowaki-etal-2019-event}.

However, previous research has not thoroughly leveraged the relation information for supervision considering the misaligned schema between human and LLMs \cite{peng2023doesincontextlearningfall} and only discussed the in-context learning (ICL) approach. Unlike the previous work, we formulate ECRE as a binary QA task with event structures to fully utilize the schema, context, and event-associated information. We also fine-tune LLMs to conduct more comprehensive analysis.

\section{Methodology}

Following the overall framework of KnowQA illustrated in Figure \ref{fig:framework}, in this section, we introduce each part of the framework, i.e., the \textit{Event Structure Construction} module and the \textit{Binary Question Answering} module, in detail.

\subsection{Problem Definition}

Following the previous research in EE \cite{du-cardie-2020-event,deng-etal-2021-ontoed}, we define our Event-Event Causal Relation Extraction (ECRE) task as two subtasks: \textbf{Event Causality Identification (ECI)}, which identifies the existence of causal relationships between event mentions, and \textbf{Causal Relation Classification (CRC)}, which classifies the event pairs containing causal relationships into their corresponding relation types. Formally, given a document $D = \{w_1, w_2, \ldots, w_N\}$ that contains multiple sentences ($N$ is the number of words in the document), we use $E = \{e_1, e_2, \ldots\}$ to denote the set of event mentions, $A = \{a_1, a_2, \ldots\}$ for event arguments, and $R = \{r_1, r_2, \ldots\}$ for the single-hop relationships of the arguments. An event mention $e_i$, an event argument $a_j$, and a relationship $r_k$ are connected if $a_j$ is an event argument of $e_i$ and $r_k$ includes $a_j$ as either a head or tail entity, and an event mention with its associated entities form an event structure. The event arguments and their single-hop relations are obtained from relevant IE models \cite{peng-etal-2023-omnievent,li-etal-2021-document,eberts-ulges-2021-end}.

Our ECRE task aims to predict both the existence of causal relationships and their corresponding relation types. Given the document $D$ and two event mentions of interest $e_h$ and $e_t$, ECI predicts whether there is a causal relationship between $e_h$ and $e_t$, and CRC predicts the specific relationship for the pair $(e_h, e_t)$, such as \texttt{Cause}, \texttt{Effect}, and \texttt{Precondition}.

\subsection{Event Structure Construction Module}

\begin{figure}[t!]
    \centering
    \includegraphics[width=1\linewidth]{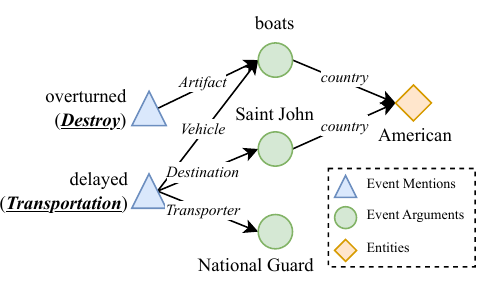}
    \caption{Event structures for the example shown in Figure \ref{fig:framework}, consisting of event mentions, event arguments, and the single-hop relationships of the arguments.}
    \label{fig:event-structure}
\end{figure}

In the first stage of KnowQA, we extract the event arguments and their single-hop relationships to form document-level event structures. We extend the definition proposed by Automatic Content Extraction (ACE), consisting of event mentions and event arguments \cite{9627684}, with the single-hop relationships of arguments to enrich their information in contexts, as examples shown in Figure \ref{fig:event-structure}. An intuitive option is to utilize LLMs for this procedure; however, since they have been found to be insufficient for IE \cite{li2023evaluatingchatgptsinformationextraction,peng2023doesincontextlearningfall}, we adopt PLM-based approaches to construct event structures, consisting of three steps: \textit{Event Detection}, \textit{Event Argument Extraction}, and \textit{Joint Entity and Relation Extraction}.

\paragraph{Event Detection.} We first conduct event detection to classify the event mentions into pre-defined  schema. To ensure the richness of classification, we adopt the KAIROS\footnote{\url{https://www.ldc.upenn.edu/collaborations/current-projects}} ontology, a superset of ACE 2005 \cite{walker2006ace} that consists of $50$ event types and $59$ argument roles\footnote{\url{https://github.com/raspberryice/gen-arg/blob/main/event_role_KAIROS.json}}, to classify the event mentions. We train an event detection model using the CLEVE \cite{wang-etal-2021-cleve} PLM on the WikiEvents dataset \cite{li-etal-2021-document} to classify the event mentions to their most likely belonged event type in the KAIROS ontology.

\paragraph{Event Argument Extraction.} Afterwards, we follow previous IE toolkits \cite{wen-etal-2021-resin,du-etal-2022-resin} to extract the event arguments with BART-Gen \cite{li-etal-2021-document}, a generative model for document-level event argument extraction (EAE). It formulates EAE as a conditional generation, consisting of the original document and a series of blank event templates with respect to the arugment roles for each event type in the KAIROS ontology, e.g., ``\textit{<arg1> damaged <arg2> using <arg3> instrument in <arg4> place}''. We adopt the templates defined by \newcite{li-etal-2021-document} to extract arguments for the event mentions.

\paragraph{Joint Entity and Relation Extraction.} Finally, we extract the single-hop relationships of the event arguments. To obtain richer relationships, we utilize a joint entity and relation extraction model named JEREX \cite{eberts-ulges-2021-end} to complete this process, which is a model pre-trained on the DocRED dataset \cite{yao-etal-2019-docred} with $6$ named entity types and $96$ relation types. After the entities and their relationships are extracted, we match the entities with the event arguments. We make revisions to the event arguments and entities with higher spans once and select the corresponding head or tail entities with the largest spans.

\subsection{Binary Question Answering Module}

Following the construction of event structures, we formulate ECRE as a binary QA task with single-turn and multi-turn QA strategies with the constructed event structures. Specifically, single-turn QA is proposed for identifying causal relationships, and multi-turn QA is for identifying and classifying the relationships, adding specific relation types in the questions.

\paragraph{Single-turn QA.} In the single-turn QA strategy, we make use of the prompt proposed by previous work \cite{man-etal-2022-event,gao-etal-2023-chatgpt} and incorporate the event structures of the two event mentions into the prompt. It starts with the word ``\textit{Input:}'',  followed by the original text and the event arguments and their relationships obtained from the IE models. Finally, we add a question designed to predict the existence of causal relationships, accompanied with the word ``\textit{Answer:}'' that ask the model to answer the binary question:

\begin{quote}
    \textit{Input: \{source\_text\}}
    
    \textit{Arguments of \{head\_evt\}: \{head\_args\}}
    
    \textit{Arguments of \{tail\_evt\}: \{tail\_args\}}
    
    \textit{Argument relationships: \{relations\}}
    
    \textit{Question: Is there a causal relationship between ``\{head\_evt\}'' and ``\{tail\_evt\}''?}
    
    \textit{Answer:}
\end{quote}
In the prompt illustrated above, ``\textit{\{head\_args\}}'' and ``\textit{\{tail\_args\}}'' are the arguments of the event mentions, which are listed in sequential following the extraction results. Generally, it appears like ($m$ is the number of arguments of the event mention):

\begin{quote}
    \textit{<Argument $\mathit{1}$>, ..., <Argument $m$>}
\end{quote}
Following \newcite{li-du-2023-leveraging}, we list the argument relationships, denoted as ``\textit{\{relations\}}'', with (subject, relation, object) triples. For example ($n$ is the number of triples of the arguments):
\begin{quote}
    \textit{(<Head $\mathit{1}$>, <Relation $\mathit{1}$>, <Tail $\mathit{1}$>)}, \\
    \textit{...}, \\
    \textit{(<Head $n$>, <Relation $n$>, <Tail $n$>)}.
\end{quote}

\paragraph{Multi-turn QA.} Considering that LLMs have misaligned schema understanding in IE tasks \cite{peng2023doesincontextlearningfall}, we construct multi-turn QA prompts based on the specific relation types and regard them as additional supervision for ECRE, for example:

\begin{quote}
    \textit{Input: \{source\_text\}}
    
    \textit{Arguments of \{head\_evt\}: \{head\_args\}}
    
    \textit{Arguments of \{tail\_evt\}: \{tail\_args\}}
    
    \textit{Argument relationships: \{relations\}}
    
    \textit{Question: Is ``\{head\_evt\}'' \{relation\_ty-pe\} ``\{tail\_evt\}''?}
    
    \textit{Answer:}
\end{quote}
In the prompt illustrated above, ``\textit{\{relation\_type\}}'' can be ``\textit{caused by}'' or ``\textit{preconditioned by}'', following the relation types (\texttt{Cause/Effect} and \texttt{Precondition}) annotated in ECRE datasets. We iterate the relation types and prompt LLMs in both directions to obtain the causal relationship between each event mention pair.

Intuitively, this QA strategy contains two potential problems: (1) the expression of causal relationship usually varies (e.g., ``\textit{cause}'' and ``\textit{caused by}''), so it is necessary to test on multiple causal expressions to ensure the generalizability of the proposed method; and (2) because we ask the model for a specific event mention pair for multiple times, it may potentially answer positively to all questions, yet the QA process terminates as long as the model receives a positive answer. As a result, we conduct additional analysis in Section \ref{sec:additional} on the impact of causal expressions and the order of the questions to model performance.

\section{Experiments}

\subsection{Datasets}

We conducted experiments on two commonly used document-level ECRE datasets: MECI \cite{lai-etal-2022-meci} and MAVEN-ERE \cite{wang-etal-2022-maven}. MECI contains the \texttt{Cause} and \texttt{Effect} relationships, and MAVEN-ERE contains the \texttt{Cause} and \texttt{Precondition} relationships. During our experiments, we selected the English subset from MECI, and we followed \newcite{gao-etal-2023-chatgpt} and \newcite{chen-etal-2024-improving-large} to sample a subset from MAVEN-ERE. The characteristics of the datasets are shown in Table \ref{tab:data-characteristics}, in which the number of event arguments and their relationships were derived from the IE models for the MECI dataset, and we utilized the golden argument annotation from MAVEN-ARG \cite{wang-etal-2024-maven} for the MAVEN-ERE dataset.

\begin{table}[t!]
    \centering
    \resizebox{0.88\linewidth}{!}{\begin{tabular}{r|cc}
        \toprule
         & \textbf{MECI} & \textbf{MAVEN-ERE} \\
        \midrule
        \#Document & $438$ & $4,480$ \\
        \#Sentence & $2,190$ & $49,873$ \\
        \#Avg. Token/Doc. & $146$ & $385$ \\
        \#Event & $8,732$ & $112,276$ \\
        \#Evt. Relation & $4,100$ & $57,992$ \\
        \midrule
        \#Argument & $11,593$ & $290,613$ \\
        \#Arg. Relation & $1,751$ & $-$ \\
        \bottomrule
    \end{tabular}}
    \caption{Characteristics of the MECI and MAVEN-ERE datasets.}
    \label{tab:data-characteristics}
\end{table}

\begin{table*}[t!]
    \centering
    \resizebox{0.92\linewidth}{!}{\begin{tabular}{l|ccc|ccc|ccc|ccc}
        \toprule
         & \multicolumn{6}{c|}{\textbf{MECI}} & \multicolumn{6}{c}{\textbf{MAVEN-ERE}} \\
        \cmidrule{2-13}
        \textbf{Model} & \multicolumn{3}{c|}{\textbf{ECI}} & \multicolumn{3}{c|}{\textbf{CRC}} & \multicolumn{3}{c|}{\textbf{ECI}} & \multicolumn{3}{c}{\textbf{CRC}} \\
        \cmidrule{2-13}
         & \textbf{P} & \textbf{R} & \textbf{F1} & \textbf{P} & \textbf{R} & \textbf{F1} & \textbf{P} & \textbf{R} & \textbf{F1} & \textbf{P} & \textbf{R} & \textbf{F1} \\
        \midrule
         & \multicolumn{12}{c}{\textbf{GPT-3.5}} \\
        \midrule
        Single-turn & $24.5$ & $\mathbf{92.3}$ & $38.8$ & $-$ & $-$ & $-$ & $25.8$ & $\underline{67.7}$ & $37.4$ & $-$ & $-$ & $-$ \\
        $\quad$ \textit{w/ Args.} & $27.3$ & $\underline{80.9}$ & $40.8$ & $-$ & $-$ & $-$ & $\underline{27.2}$ & $\mathbf{73.8}$ & $\mathbf{39.8}$ & $-$ & $-$ & $-$ \\
        $\quad$ \textit{w/ Rels.} & $28.8$ & $72.0$ & $41.2$ & $-$ & $-$ & $-$ & $-$ & $-$ & $-$ & $-$ & $-$ & $-$ \\
        \midrule
        Multi-turn & $32.8$ & $76.3$ & $\mathbf{45.9}$ & $20.9$ & $\mathbf{48.7}$ & $29.3$ & $27.1$ & $55.0$ & $36.3$ & $\mathbf{15.0}$ & $\underline{18.6}$ & $\underline{16.6}$ \\
        $\quad$ \textit{w/ Args.} & $\underline{33.2}$ & $64.2$ & $43.8$ & $\underline{24.1}$ & $\underline{46.8}$ & $\underline{31.9}$ & $\mathbf{27.7}$ & $64.4$ & $\underline{38.8}$ & $\underline{13.9}$ & $\mathbf{23.9}$ & $\mathbf{17.6}$ \\
        $\quad$ \textit{w/ Rels.} & $\mathbf{34.9}$ & $59.1$ & $\underline{43.9}$ & $\mathbf{25.5}$ & $43.3$ & $\mathbf{32.1}$ & $-$ & $-$ & $-$ & $-$ & $-$ & $-$ \\
        \midrule
         & \multicolumn{12}{c}{\textbf{$\textbf{Flan-T5}_\textit{\textbf{XL}}$}} \\
        \midrule
        Single-turn & $30.2$ & $\underline{83.5}$ & $44.4$ & $-$ & $-$ & $-$ & $26.0$ & $57.9$ & $35.9$ & $-$ & $-$ & $-$ \\
        $\quad$ \textit{w/ Args.} & $30.4$ & $\mathbf{83.9}$ & $44.6$ & $-$ & $-$ & $-$ & $26.4$ & $\mathbf{66.7}$ & $\underline{37.8}$ & $-$ & $-$ & $-$ \\
        $\quad$ \textit{w/ Rels.} & $31.7$ & $83.2$ & $45.9$ & $-$ & $-$ & $-$ & $-$ & $-$ & $-$ & $-$ & $-$ & $-$ \\
        \midrule
        Multi-turn & $\mathbf{40.3}$ & $64.8$ & $\underline{49.7}$ & $\mathbf{35.4}$ & $56.9$ & $\mathbf{43.6}$ & $\underline{28.7}$ & $52.2$ & $37.0$ & $\underline{17.4}$ & $\underline{38.4}$ & $\underline{24.0}$ \\
        $\quad$ \textit{w/ Args.} & $\underline{39.0}$ & $66.9$ & $49.3$ & $\underline{34.0}$ & $\underline{58.3}$ & $42.9$ & $\mathbf{29.2}$ & $\underline{62.1}$ & $\mathbf{39.7}$ & $\mathbf{18.4}$ & $\mathbf{45.4}$ & $\mathbf{26.2}$ \\
        $\quad$ \textit{w/ Rels.} & $38.8$ & $71.8$ & $\mathbf{50.4}$ & $33.5$ & $\mathbf{62.0}$ & $\underline{43.5}$ & $-$ & $-$ & $-$ & $-$ & $-$ & $-$ \\
        \bottomrule
    \end{tabular}}
    \caption{Performance of KnowQA against baselines on the MECI and MAVEN-ERE datasets under the \textit{zero-shot} setting. The best and second-best results for each model are highlighted in \textbf{bold} and \underline{underlined}, respectively.}
    \label{tab:zero-shot}
\end{table*}

\subsection{Baselines}

We compared the performance of KnowQA against the following state-of-the-art baselines from existing ECRE research:
(1) \textbf{PLM} \cite{tran-phu-nguyen-2021-graph} classifies causal relationships after obtaining event representations;
(2) \textbf{Know} \cite{ijcai2020p499} retrieves related concepts and relations for event mentions from ConceptNet to augment input texts;
(3) \textbf{RichGCN} \cite{tran-phu-nguyen-2021-graph} enriches the event representations by constructing interaction graphs between essential objects;
(4) \textbf{ERGO} \cite{chen-etal-2022-ergo} builds event relational graphs to convert ECRE into a node classification problem;
(5) \textbf{DiffusECI} \cite{Man_Dernoncourt_Nguyen_2024} develops a diffusion model to generate causal label representations to eliminate irrelevant components;
(6) \textbf{HOTECI} \cite{man-etal-2024-hierarchical} leverages optimal transport to select the most important words and sentences from full documents;
(7) \textbf{GIMC} \cite{he-etal-2024-zero-shot} constructs a heterogeneous graph interaction network to model long-distance dependencies between events.
We followed the original implementations of the baselines using the XLM-RoBERTa PLM \cite{conneau-etal-2020-unsupervised}.

\subsection{Experimental Setup}

We conducted experiments under both zero-shot and fine-tuning settings with two LLMs: GPT-3.5 and Flan-T5 \cite{JMLR:v25:23-0870}. Specifically, we experimented with GPT-3.5 (\texttt{gpt-3.5-turbo-} \texttt{0125}) under the zero-shot setting from its official API\footnote{\url{https://platform.openai.com/}}, and we set the temperature as $0$ to stabalize the outputs. We also conducted experiments using $\text{Flan-T5}_\textit{XL}$ under zero-shot and $\text{Flan-T5}_\textit{Large}$ under fine-tuning settings. During the fine-tuning process, we set the batch size as $4$, the gradient accummulation steps as $4$, the learning rate as $5e-5$, and the number of epochs as $5$, and selected the best validation models to test performance on the test set. The main experiments were conducted on a single GeForce RTX 3090 graphic card.
Detailed experimental settings for the event structure construction process are organized in Appendix \ref{sec:event-structure-exp}.

\subsection{Main Results}

\begin{table}[t!]
    \centering
    \resizebox{\linewidth}{!}{\begin{tabular}{l|ccc|ccc}
        \toprule
        \multirow{2}{*}{\textbf{Model}} & \multicolumn{3}{c|}{\textbf{ECI}} & \multicolumn{3}{c}{\textbf{CRC}} \\
        \cmidrule{2-7}
         & \textbf{P} & \textbf{R} & \textbf{F1} & \textbf{P} & \textbf{R} & \textbf{F1} \\
        \midrule
        PLM$^{[\diamond]}$ & $56.4$ & $77.8$ & $65.4$ & $45.9$ & $59.7$ & $51.9$ \\
        Know$^{[\diamond]}$ & $42.4$ & $75.7$ & $54.3$ & $34.3$ & $47.2$ & $39.7$ \\
        RichGCN$^{[\diamond]}$ & $63.5$ & $\mathbf{79.2}$ & $70.5$ & $52.5$ & $63.6$ & $57.5$ \\
        DiffusECI & $70.1$ & $68.3$ & $69.2$ & $-$ & $-$ & $-$ \\
        HOTECI & $66.6$ & $67.1$ & $66.8$ & $-$ & $-$ & $-$ \\
        ERGO & $-$ & $-$ & $-$ & $55.0$ & $57.5$ & $56.2$ \\
        GIMC & $-$ & $-$ & $-$ & $61.5$ & $58.4$ & $59.9$ \\
        \midrule
         & \multicolumn{6}{c}{$\textbf{Flan-T5}_{\textbf{\textit{Large}}}$} \\
        \midrule
        Single-turn & $67.3$ & $75.7$ & $71.2$ & $-$ & $-$ & $-$ \\
        $\quad$ \textit{w/ Args.} & $69.1$ & $75.4$ & $72.1$ & $-$ & $-$ & $-$ \\
        $\quad$ \textit{w/ Rels.} & $70.2$ & $\underline{78.1}$ & $\mathbf{73.9}$ & $-$ & $-$ & $-$ \\
        \midrule
        Multi-turn & $65.5$ & $71.8$ & $68.5$ & $59.6$ & $65.3$ & $62.3$ \\
        $\quad$ \textit{w/ Args.} & $\mathbf{71.2}$ & $75.0$ & $73.0$ & $\mathbf{64.3}$ & $\underline{67.7}$ & $\mathbf{66.0}$ \\
        $\quad$ \textit{w/ Rels.} & $\underline{70.8}$ & $76.3$ & $\underline{73.5}$ & $\underline{63.2}$ & $\mathbf{68.1}$ & $\underline{65.6}$ \\
        \bottomrule
    \end{tabular}}
    \caption{Performance of KnowQA against baselines on the MECI dataset under the \textit{fine-tuning} setting. The best and second-best results are highlighted in \textbf{bold} and \underline{underlined}, respectively. $^{[\diamond]}$ denotes that the results are from our reproduction.}
    \label{tab:fine-tuning}
\end{table}

The main experimental results of KnowQA against baselines under zero-shot and fine-tuning settings are presented in Tables \ref{tab:zero-shot} and \ref{tab:fine-tuning}, respectively. From the tables, we have the following observations:

Firstly, under the zero-shot setting, both GPT-3.5 and $\text{Flan-T5}_{\textit{XL}}$ fell short on the ECI task no matter the identification and classification of causal relationships and exhibited high-level causal hallucination issues, which tended to assume the existence of causal relationships between event mentions. It can be observed by the unsatisfied overall performance and the low precision and high recall illustrated in the tables. $\text{Flan-T5}_{\textit{XL}}$ performed much better than GPT-3.5, whose performance was close to the fine-tuned Know baseline; it also achieved better performance on the classification of causal relationships, indicating its better reasoning ability compared with GPT-3.5.

Secondly, after fine-tuning $\text{Flan-T5}_\textit{Large}$, it outperformed all baselines and achieved state-of-the-art on both the identification and classification of causal relationships, even though its parameter is smaller than $\text{Flan-T5}_\textit{XL}$. Besides, the event structures were notably helpful in making better ECRE predictions, and the performance under both zero-shot and fine-tuning settings was correlated with the completeness of event structures, i.e., without event structures $<$ with event arguments $<$ with event arguments and their relationships.
Compared with MECI, the event structures were more helpful for MAVEN-ERE, whose model performance, after incorporating them, consistently outperformed the performance without event structures.
Nevertheless, the experimental results illustrated in the tables were enough to indicate the high transferability between different IE tasks and the effectiveness of rich information for generative models when conducting reasoning tasks. 

Finally, multi-turn QA was more effective in both identifying and classifying causal relationships than single-turn QA under the zero-shot setting and was able to alleviate the causal hallucination issues revealed in LLMs by observing the decreased recall and increased precision values; however, after fine-tuning the LLMs, the multi-turn QA strategy was adept for the classification, while the single-turn QA strategy was effective for the identification of causal relationships. By comparing differences between precision and recall values, it is evident that the differences between them significantly decreased after fine-tuning the models, indicating that the causal hallucination issues reported by LLMs could be solved after fine-tuning them. In this case, multi-turn QA detected more event mention pairs that do not contain causal relationships, but it would also omit more pairs that contain relationships compared with the single-turn QA strategy. The false negative predictions of the questions had a high likelihood of accumulating and decreasing the prediction results when identifying the existence of causal relationships.

\subsection{Effectiveness of Event Structures}

From the experimental results illustrated in Tables \ref{tab:zero-shot} and \ref{tab:fine-tuning}, it is evident that the event structures are helpful for LLMs to make better ECRE predictions by comparing the results with respect to different completeness of event structures. We analyzed the intra-sentence and inter-sentence performance of the models and conducted a case study to further understand the efficacy of event structures.

\begin{table}[t!]
    \centering
    \resizebox{\linewidth}{!}{\begin{tabular}{l|cc|cc}
        \toprule
        \multirow{2}{*}{\textbf{Model}} & \multicolumn{2}{c|}{\textbf{ECI}} & \multicolumn{2}{c}{\textbf{CRC}} \\
        \cmidrule{2-5}
         & \textbf{Intra F1} & \textbf{Inter F1} & \textbf{Intra F1} & \textbf{Inter F1} \\
        \midrule
        Single-turn & $\underline{77.8}$ & $12.5$ & $-$ & $-$ \\
        $\quad$ \textit{w/ Args.} & $76.0$ & $\mathbf{36.0}^*$ & $-$ & $-$ \\
        $\quad$ \textit{w/ Rels.} & $\mathbf{78.7}$ & $\underline{34.3}$ & $-$ & $-$ \\
        \midrule
        Multi-turn & $74.2$ & $27.4$ & $67.5$ & $25.3$ \\
        $\quad$ \textit{w/ Args.} & $\underline{76.8}$ & $\underline{31.5}$ & $\mathbf{69.4}$ & $\underline{28.3}$ \\
        $\quad$ \textit{w/ Rels.} & $\mathbf{77.0}$ & $\mathbf{40.0}^*$ & $\underline{68.9}$ & $\mathbf{34.7}^*$ \\
        \bottomrule
    \end{tabular}}
    \caption{Intra-sentence and inter-sentence performance of ECRE on the MECI dataset with the $\text{Flan-T5}_\textit{Large}$ model. $^*$ denotes statistical significance ($p < 0.01$).}
    \label{tab:intra-inter}
\end{table}

\paragraph{Intra- and Inter-sentence Performance.} Table \ref{tab:intra-inter} presents the intra-sentence and inter-sentence performance after fine-tuning the $\text{Flan-T5}_\textit{Large}$ model. From the table, we observed that the original $\text{Flan-T5}_\textit{Large}$ was not proficient in inter-sentence ECRE (e.g., an F1-score of $12.5$ in ECI under the single-turn QA strategy) and had a large gap with its intra-sentence performance. However, after incorporating the event structures, while the improvement of the intra-sentence performance was minor, the inter-sentence performance increased by a significant margin, and its gap with intra-sentence performance also decreased significantly. This indicates the effectiveness of document-level event structure in document-level ECRE predictions.

\begin{table}[t!]
    \centering
    \small
    \begin{tabular}{p{0.95\linewidth}}
        \toprule
        \textbf{Input:} Once that \textbf{\color{NavyBlue} \textit{happened}}, the INS mode would change from ``armed'' to ``capture'' and the \textbf{\color{PineGreen} \textit{plane}} would \textbf{\color{NavyBlue} \textit{track}} the flight-planned course from then on. The HEADING mode of the \textbf{\color{PineGreen} \textit{autopilot}} would normally be \textbf{\color{BurntOrange} \textit{engaged}} sometime after take off to comply with vectors from \textbf{\color{PineGreen} \textit{ATC}}, and then after \textbf{\color{BurntOrange} \textit{receiving}} appropriate ATC clearance, to guide the plane to intercept the desired INS course line. \\
        \midrule
        \textbf{Event Mention Pair:} (\textbf{\color{NavyBlue} \textit{happened}}, \textbf{\color{NavyBlue} \textit{track}}) \\
        \textbf{Argument(s) of \textit{\color{NavyBlue} happened}:} (None) \\
        \textbf{Argument(s) of \textit{\color{NavyBlue} track}:} \textbf{\color{PineGreen} \textit{plane}} \\
        \textbf{Prediction without Event Structure:} \verb|None| \\
        \textbf{Prediction with Event Structure:} \verb|Cause| \\
        \midrule
        \textbf{Event Mention Pair:} (\textbf{\color{BurntOrange} \textit{receiving}}, \textbf{\color{BurntOrange} \textit{engaged}}) \\
        \textbf{Argument(s) of \textit{\color{BurntOrange} receiving}:} \textbf{\color{PineGreen} \textit{ATC}} \\
        \textbf{Argument(s) of \textit{\color{BurntOrange} engaged}:} \textbf{\color{PineGreen} \textit{autopilot}} \\
        \textbf{Prediction without Event Structure:} \verb|Cause| \\
        \textbf{Prediction with Event Structure:} \verb|Effect| \\
        \bottomrule
    \end{tabular}
    \caption{Examples of the event mention pairs that can be correctly classified with event structures (\textbf{\color{NavyBlue} \textit{Blue}}: event mentions of the first example, \textbf{\color{BurntOrange} \textit{Orange}}: event mentions of the second example, \textbf{\color{PineGreen} \textit{Green}}: event arguments extracted by IE models).}
    \label{tab:case}
\end{table}

\paragraph{Case Study.}

We sampled $50$ cases that could not derive correct prediction without event structures but were able to predict correctly with them. We categorized them into two cases: \textit{identifying implicit causal relationships} and \textit{correcting mispredictions}. Table \ref{tab:case} illustrates examples concerning the two cases.

In the first example, the original model could not identify the causal relationship because it was not explicitly expressed, i.e., there was not explicit causal clues, such as ``\textit{cause}'' and ``\textit{because}'', between the event mentions; however, with the event argument ``plane'' of the event mention ``track'', the model was able to detect the ``\texttt{Cause}'' relationship between the event mentions. In the second example, the original model classified the relationship as ``\texttt{Cause}'' because there was a confusable temporal clue between the event mentions, i.e., ``\textit{after}''; however, the temporal clue was for the events ``take off'' but not ``receiving''. With the event structures, the model could predict correct relationships between them, i.e., engaged $\xrightarrow{cause}$ receiving, by potentially leveraging the relationships between the two event arguments, e.g., the engagement of autopilot may affect ATC (air traffic control) in some cases.

\subsection{Additional Analysis}
\label{sec:additional}

\paragraph{Effects of Causal Expressions.}

\begin{figure}[t!]
    \centering
    \includegraphics[width=1\linewidth]{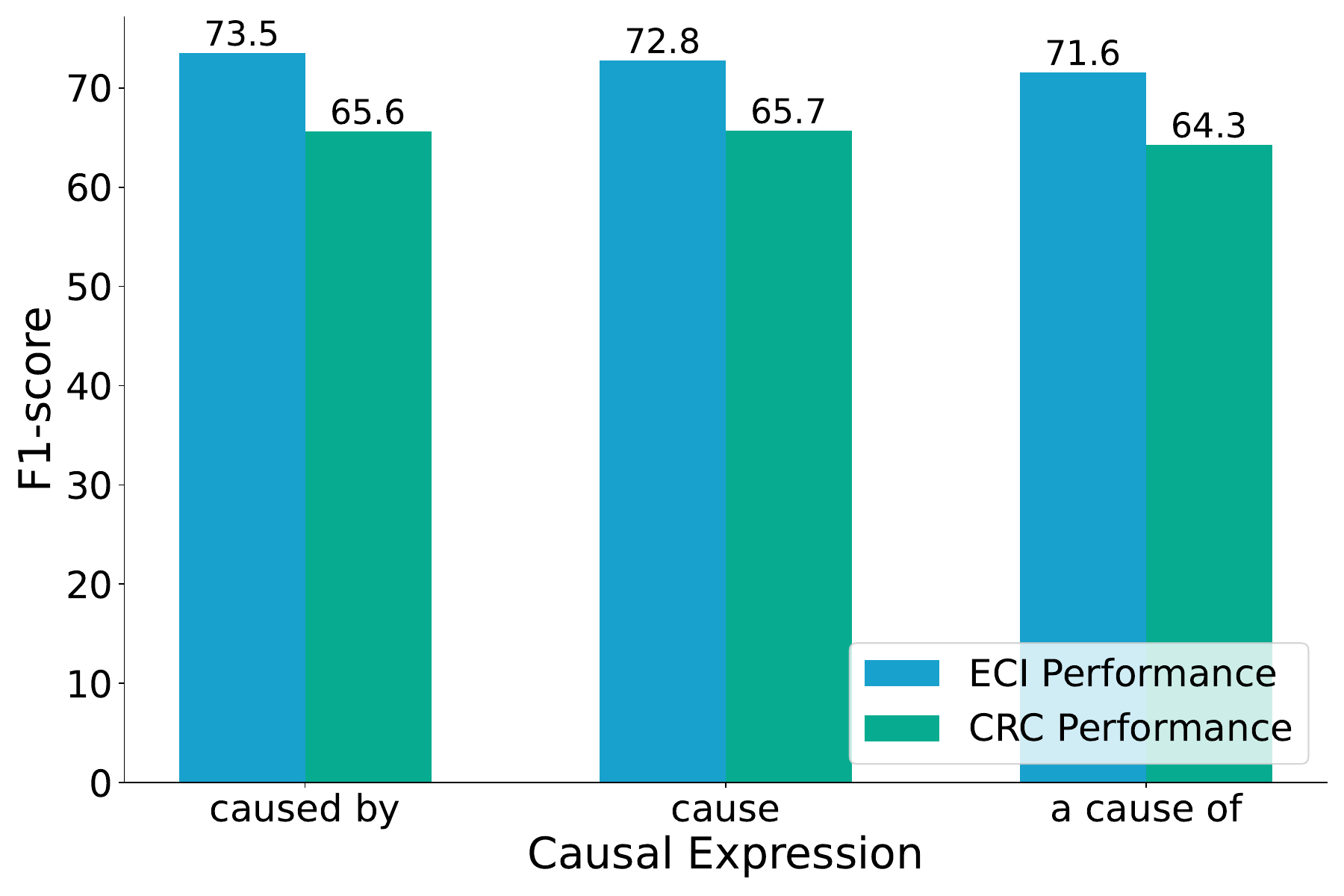}
    \caption{Performance of $\text{Flan-T5}_\textit{Large}$ with complete event structures using different causal expressions on the MECI dataset.}
    \label{fig:reverse}
\end{figure}

In the main experiments, we formulated our questions with passive voice, i.e.,  ``\textit{caused by}'' and ``\textit{preconditioned by}''. Because the expression of causal relationship usually varies, for a pair of event mentions $(e_h, e_t)$, we conducted additional experiments with two different causal expressions, including (1) ``\textit{Does $\mathit{e_h}$ cause $\mathit{e_t}$?}'' and (2) ``\textit{Is $\mathit{e_h}$ a cause of $\mathit{e_t}$?}'' As shown in Figure \ref{fig:reverse}, our method achieved similar performance when using different causal expressions, which all outperformed the baseline models, indicating our high generalizability; however, the performance using the expression ``\textit{Is $\mathit{e_t}$ caused by $\mathit{e_h}$?}'' achieved the best result, indicating that LLMs are more favored in the passive voice when understanding causal relationships.

\begin{figure}[t!]
    \centering
    \includegraphics[width=1\linewidth]{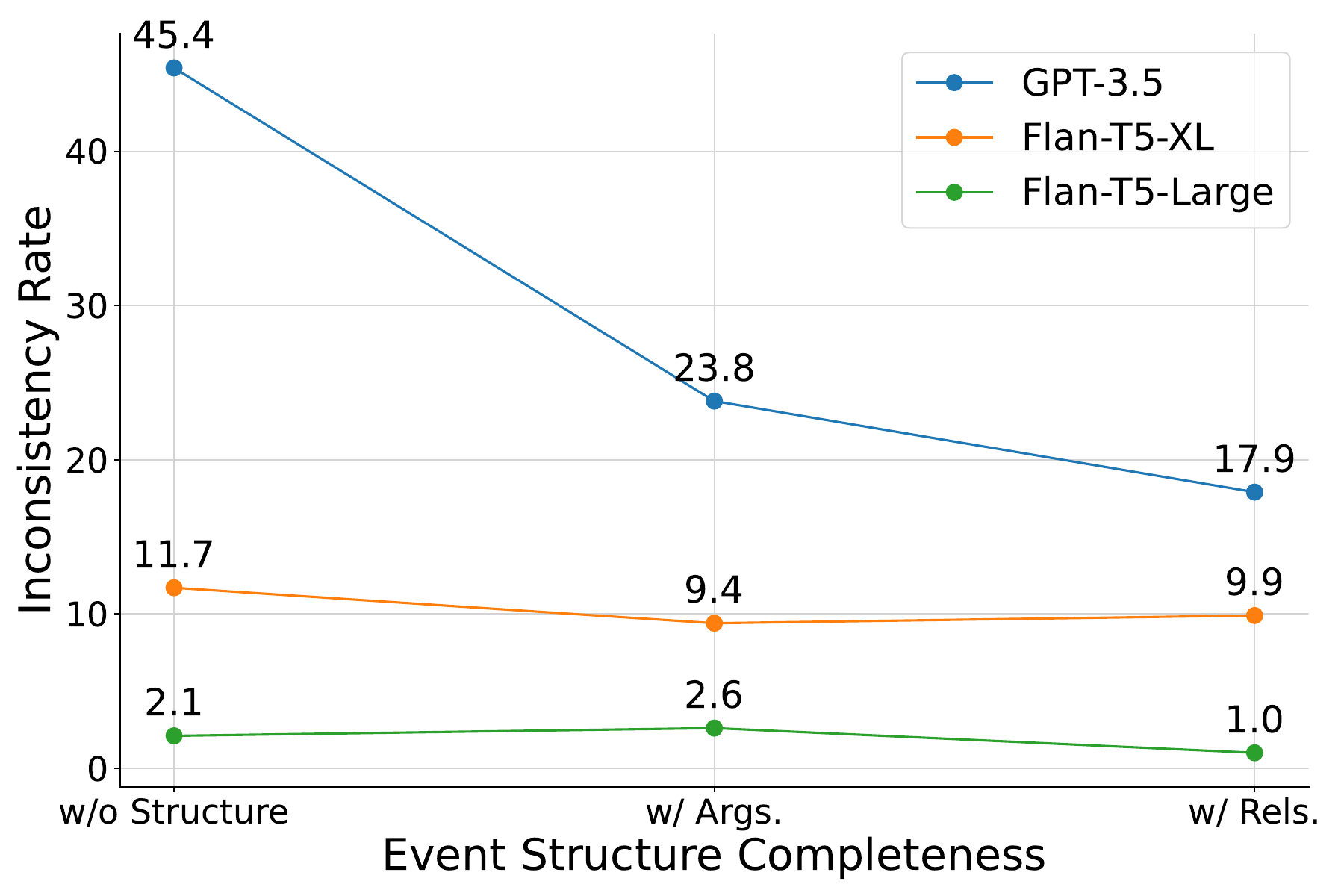}
    \caption{Inconsistency evaluation results of the models on the MECI dataset.}
    \label{fig:inconsistency}
\end{figure}

\paragraph{Inconsistency Evaluation.}

To guarantee that the order of the questions does not significantly alter model predictions under the multi-turn QA strategy, we defined an additional evaluation metric called ``inconsistency'' to evaluate the models. Specifically, we inferred the models with both binary questions with reversed causal directions (e.g., \texttt{Cause} and \texttt{Effect}) and obtained model predictions. We formulated the ``inconsistency'' metric as the proportion of the number of event pairs that the model predicts both questions positive to the number of pairs that the model predicts at least one positive, formally:
\begin{equation}
    \text{Inconsistency} = \frac{\text{\# of Both Positive}}{\text{\# of At Least One Positive}}.
\end{equation}

Figure \ref{fig:inconsistency} illustrates the results of the inconsistency of the experimented models. It can be observed that the models under the zero-shot setting, particularly GPT-3.5, exhibited a high inconsistency problem. This well explains the reason why the performance difference of GPT-3.5's comprehensible result in the identification of causal relationships but unsatisfactory in classification compared with $\text{Flan-T5}_\textit{XL}$: the model had a high likelihood of generating a positive response in the first turn and predicting the relationship as \texttt{Effect}.
Under the fine-tuning setting, $\text{Flan-T5}_\textit{Large}$ had a minimal inconsistency. Notably, the incorporation of event structures could alleviate the inconsistency problem on all models, and the inconsistency rate of $\text{Flan-T5}_\textit{Large}$ with complete event structures was only $1.0$, indicating that the order of the questions hardly affects the model prediction and the high reliability of our proposed method.

\section{Conclusion and Future Work}

We introduced KnowQA, a novel method for ECRE that formulates the task as a binary QA task with the utilization of cross-task knowledge in IE, i.e., document-level event structures, consisting of two stages: \textit{Event Structure Construction} and \textit{Binary Question Answering}. Experimental results on the MECI and MAVEN-ERE datasets demonstrated that the effectiveness, high generalizability, and low inconsistency of KnowQA, particularly with complete event structures after fine-tuning the models. In the future, we will test our method with more event relations (e.g., temporal and subevent relations) and more languages to further validate the generalizability of our proposed method.

\section*{Limitations}

The limitations of KnowQA in the current work are as follows: (1) The event structures were constructed with relevant IE models on ECRE datasets, which were not golden labels. Although they have been proven to be helpful for the ECRE task, the errors from the event structure construction process may have negative impacts on ECRE predictions. (2) We only tested KnowQA with limited LLMs and only on English corpus. Future investigations of KnowQA with other commonly used LLMs (e.g., Llama and Mistral) and languages (e.g., Danish and Spanish) can be conducted in the future to validate the generalizability of KnowQA in more scenarios.


\bibliography{custom}

\appendix

\section{Experimental Setup for Event Structure Construction}
\label{sec:event-structure-exp}

\paragraph{Event Detection} We trained an event detection using the OmniEvent toolkit \cite{peng-etal-2023-omnievent} on the WikiEvents dataset \cite{li-etal-2021-document}, and we selected CLEVE \cite{wang-etal-2021-cleve} as the PLM. During the training process, we followed the original experimental settings proposed in OmniEvent\footnote{\url{https://github.com/THU-KEG/OmniEvent/blob/main/config/all-models/ed/tc/roberta-large/cleve.yaml}}. We set the batch size as $40$, the learning rate as $1e-5$, and the number of epochs as $30$, and we selected the best validation model on the WikiEvents dataset to conduct event detection on our datasets.

\paragraph{Event Argument Extraction \& Joint Entity and Relation Extraction} We adopted the pre-trained models released by BART-Gen\footnote{\url{https://github.com/raspberryice/gen-arg}} \cite{li-etal-2021-document} and JEREX\footnote{\url{https://github.com/lavis-nlp/jerex}} \cite{eberts-ulges-2021-end} to conduct EAE and joint entity and relation extraction on our datasets, separately.

\end{document}